\documentclass[sigconf]{acmart}

\usepackage{multirow} 
\usepackage{subcaption}
\usepackage{enumitem}
\usepackage{booktabs, tabularx}
\usepackage{xcolor}
\usepackage{makecell} 
\usepackage{xurl}
\usepackage{arydshln}
\usepackage{tcolorbox}
\usepackage[symbol]{footmisc}

\newcommand{\new}[1]{\textcolor{black}{#1}}

\AtBeginDocument{
  }

\copyrightyear{2025}
\acmYear{2025}
\setcopyright{cc}
\setcctype{by}
\acmConference[SIGIR-AP 2025]{Proceedings of the 2025 Annual International ACM SIGIR Conference on Research and Development in Information Retrieval in the Asia Pacific Region}{December 7--10, 2025}{Xi'an, China}
\acmBooktitle{Proceedings of the 2025 Annual International ACM SIGIR Conference on Research and Development in Information Retrieval in the Asia Pacific Region (SIGIR-AP 2025), December 7--10, 2025, Xi'an, China}\acmDOI{10.1145/3767695.3769488}
\acmISBN{979-8-4007-2218-9/2025/12}

\begin{document}

\title{Beyond GeneGPT: A Multi-Agent Architecture with Open-Source LLMs for Enhanced Genomic Question Answering}


\author{Haodong Chen}
\affiliation{
  \institution{The University of Queensland}
  \city{Brisbane}
  \country{Australia}
}
\email{haodong.chen1@student.uq.edu.au}

\author{Guido Zuccon}
\affiliation{%
  \institution{The University of Queensland}
  \city{Brisbane}
  \country{Australia}
}
\email{g.zuccon@uq.edu.au}

\author{Teerapong Leelanupab}
\affiliation{%
	\institution{The University of Queensland}
	\city{Brisbane}
	\country{Australia}
}
\email{t.leelanupab@uq.edu.au}

\begin{abstract}
Genomic question answering often requires complex reasoning and integration across diverse biomedical sources. GeneGPT addressed this challenge by combining domain-specific APIs with OpenAI's \textit{code-davinci-002} large language model to enable natural language interaction with genomic databases. However, its reliance on a proprietary model limits scalability, increases operational costs, and raises concerns about data privacy and generalization.



In this work, we revisit and reproduce GeneGPT in a pilot study using open source models, including Llama 3.1, Qwen2.5, and Qwen2.5 Coder, within a monolithic architecture; this allows us to identify the limitations of this approach. Building on this foundation, we then develop \textit{OpenBioLLM}, a modular multi-agent framework that extends GeneGPT by introducing agent specialization for tool routing, query generation, and response validation. This enables coordinated reasoning and role-based task execution.


OpenBioLLM matches or outperforms GeneGPT on over 90\% of the benchmark tasks, achieving average scores of 0.849 on GeneTuring and 0.830 on GeneHop, while using smaller open-source models without additional fine-tuning or tool-specific pretraining. OpenBioLLM's modular multi-agent design reduces latency \new{by 40–50\% across benchmark tasks, significantly improving efficiency without compromising model capability.} The results of our comprehensive evaluation highlight the potential of open-source multi-agent systems for genomic question answering. Code and resources are available at \url{https://github.com/ielab/OpenBioLLM}.

\end{abstract}

\ccsdesc[500]{Information systems~Specialized information retrieval}

\keywords{Genomic Question Answering, LLM, NCBI}
\maketitle
\vspace{-0.2cm}
\section{Introduction and Related Work}
\label{intro}

Large Language Models (LLMs) have achieved remarkable success in general-domain NLP tasks as well as in specialized domains, such as finance~\cite{Wu:2023:Bloomber}, law~\cite{Yue:2024:Event-Gr}, clinical medicine~\cite{Singhal:2022:Large-La}, and biomedicine~\cite{Luo:2023:BioMedGP}. In genomics, LLMs have been applied to question answering (QA) tasks, which involve retrieving accurate information about genes, variants, and sequences in response to natural language queries. Prior to GeneGPT~\cite{Jin:2024:GeneGPT:}, most approaches used general-purpose or biomedical LLMs such as BioGPT~\cite{luo2022biogpt} and BioMedLM~\cite{bolton2024biomedlm}, which generated answers solely from pretraining knowledge and often produced hallucinations on entity- or sequence-level queries. To address these limitations, GeneGPT~\cite{Jin:2024:GeneGPT:} pioneered the integration of LLMs with domain-specific APIs from the National Center for Biotechnology Information (NCBI)~\cite{Sayers:2019:Database}, enabling tool-augmented genomic question answering (QA) with improved factual accuracy.



Despite its contributions, GeneGPT is built on a monolithic pipeline using a single proprietary Codex model (\textit{code-davinci-002}), which is now deprecated and unavailable to the public. This limits GeneGPT's usability, reproducibility, and extensibility. Additionally, its single-agent architecture lacks the flexibility to support multi-hop reasoning and complex task decomposition, as required by genomic QA tasks exemplified by the GeneHop benchmark~\cite{Jin:2024:GeneGPT:}.

In this work, we revisit and reimplement GeneGPT using open-source models, including Llama 3.1, Qwen 2.5, and Qwen2.5 Coder, within its original monolithic design. Building on this foundation and examining its limitations, we propose \textit{OpenBioLLM}, a modular multi-agent framework that introduces agent specialisation for tool routing, query generation, and response validation. This architecture supports coordinated reasoning, parallel execution, and role-based task delegation to handle diverse genomic queries more effectively.

We comprehensively evaluate OpenBioLLM on twelve benchmark tasks, showing that it matches or outperforms GeneGPT on eleven, achieving average scores of 0.849 on GeneTuring and 0.830 on GeneHop, while using smaller open-source models without additional fine-tuning or tool-specific pretraining. Moreover, its modular multi-agent design reduces latency and improves interpretability.

\new{Beyond validating prior work~\cite{Jin:2024:GeneGPT:}, our contribution is the development of a robust open-source system that prioritises reproducibility, optimisation, traceability, and efficiency. Through rigorous engineering and comprehensive evaluation, we provide empirical insights into multi-agent architecture, role allocation, and error patterns, offering immediate practical benefits for genomic QA researchers and broader implications for the IR and NLP communities. Our key contributions are summarised as follows:}



\begin{itemize}[noitemsep, topsep=0.5em, leftmargin=*]

\item[1.] \textbf{Generalizability and Reproducibility:} We adapt GeneGPT to open-source LLMs, enabling reproduction without reliance on proprietary models.

\item[2.] \textbf{Improved Multi-Step Reasoning:} We enhance performance on complex tasks (e.g., GeneHop) by analyzing failure cases and leveraging complementary model strengths through modular design.

\item[3.] \textbf{Tool Use with Open-Source Models:} We show that open-source LLMs can support tool-augmented QA via prompt design and agent coordination, without requiring fine-tuning or task-specific supervision.

\item[4.] \textbf{Multi-Agent Architecture:} We introduce a modular framework with role-specialised agents for tool routing, query generation, and response validation, overcoming the rigidity of GeneGPT’s monolithic design and aligning with the growing shift toward agent-based LLM systems.



\item[5.]\textbf{Empirical Insights into Model Scaling:} \new{We find that smaller, role-faithful agents can outperform larger models that rely on reasoning shortcuts, highlighting the importance of role-fit over raw scale in agent design.}


\item[6.] \textbf{API Dependency Analysis:} We evaluate the limitations of relying on domain-specific APIs (e.g., NCBI) and provide insights for improving robustness and generalisation across biomedical QA systems.

\end{itemize}

\begin{table}[t!]
	\centering
	\vspace{-0.20cm}
	\caption{GeneTuring Tasks, Categories, Inputs, and Outputs}
	\label{tab:geneturing_inNout}
	\vspace{-0.25cm}
	\resizebox{\linewidth}{!}{%
		\begin{tabular}{lll}
			\toprule
			\textbf{GeneTuring Task} & \textbf{Input} & \textbf{Output} \\
			\midrule
			\textbf{Nomenclature} & & \\
			Gene alias & Gene alias & Official gene symbol \\
			Gene name conversion & Ensembl Gene ID & Official gene symbol \\
			\midrule
			\textbf{Genomic Location} & & \\
			Gene location & Gene name & Chromosomal location \\
			SNP location & SNP ID & Chromosomal location \\
			Gene SNP association & SNP ID & Related genes \\
			\midrule
			\textbf{Functional Analysis} & & \\
			Gene disease association & Disease name & Related genes \\
			Protein-coding genes & Gene name & Indicates if it's a protein-coding gene \\
			\midrule
			\textbf{Sequence Alignment} & & \\
			DNA to human genome & DNA sequence & Chromosomal location \\
			DNA to multiple species & DNA sequence & Chromosomal locations across species \\
			\bottomrule
		\end{tabular}%
	}
  \vspace{-12pt}
\end{table}

\section{Genomic QA Tasks}

We evaluate GeneGPT and OpenBioLLM on two genomic QA benchmarks: GeneTuring and GeneHop. These tasks assess the ability of LLMs to answer structured and multi-step questions using genomic data and APIs.

\vspace{-0.10cm}
\subsection{GeneTuring}

GeneTuring consists of twelve tasks (50 questions each)~\cite{Hou:2025:Benchmar}. Following GeneGPT~\cite{Jin:2024:GeneGPT:}, we focus on the nine tasks that interact with NCBI APIs. These tasks fall into four functional categories, as outlined in Table~\ref{tab:geneturing_inNout}, which summarizes the input and output for each:

\begin{itemize}[leftmargin=*, itemsep=0pt, topsep=0pt]
    \item \textbf{Nomenclature:} Standardizes gene names through alias identification and Ensembl-to-symbol conversion.
    \item \textbf{Genomic Location:} Retrieves chromosomal positions of genes or SNPs, as well as their associations.
    \item \textbf{Functional Analysis:} Identifies disease-associated or protein-coding genes.
    \item \textbf{Sequence Alignment:} Maps DNA sequences to genomic coordinates or identifies the species (e.g., “worm,” “chicken,” “rat”) to which the sequence belongs.
\end{itemize}
All tasks follow a single-hop QA format, with answers retrieved through structured calls to NCBI APIs (e.g., E-utils, BLAST).

\begin{table}[t]
	\centering
	\caption{GeneHop Tasks, Inputs, Outputs, and Required APIs}
	\label{tab:genehop_inNout}
	\vspace{-0.2cm}
	\resizebox{\linewidth}{!}{%
	\begin{tabular}{lll|l}
\hline
			\textbf{GeneHop Task} & \textbf{Input} &  \textbf{Output}  & \textbf{Required APIs} \\
\hline
			SNP gene function & SNP ID & Function description and associated gene & e-utils \\
\hline
			Disease gene location & Disease name & Chromosomal location & e-utils \\
\hline
			Sequence gene alias & DNA sequence & Gene alias & blast + e-utils \\
\hline
		\end{tabular}%
	}
	\vspace{-1em}
\end{table}

\subsection{GeneHop}

GeneHop evaluates a model’s ability to perform multi-hop reasoning via three tasks requiring sequential API calls and intermediate logic~\cite{Jin:2024:GeneGPT:}. Each task begins with a primary question and involves decomposing the query into multiple steps.

\begin{itemize}[leftmargin=*, itemsep=0pt, topsep=0pt]
    \item \textbf{SNP Gene Function:} Identify the gene associated with a given SNP, then retrieve its function (SNP → gene → function).
    \item \textbf{Disease Gene Location:} Retrieve the OMIM ID for a disease, identify related genes, and determine their chromosomal locations (disease → OMIM ID → gene → location).
    \item \textbf{Sequence Gene Alias:} From a DNA sequence, infer the genomic location, retrieve the gene, and return its aliases (sequence → location → gene → aliases).
\end{itemize}

\noindent These tasks simulate real-world genomic queries and test a model's ability to integrate multiple API responses while maintaining reasoning consistency. Table~\ref{tab:genehop_inNout} summarizes task inputs, outputs, and associated NCBI APIs.


\vspace{-0.10cm}
\section{GeneGPT}

\subsubsection*{GeneGPT Inference and NCBI APIs}

GeneGPT~\cite{Jin:2024:GeneGPT:} enables LLMs to interact with biomedical databases via NCBI Web APIs~\cite{Sayers:2019:Database}, combining structured prompts with tool-augmented generation to answer genomic questions.

\begin{figure}[t]
	\centering
	\vspace{-0.4cm}
	\includegraphics[width=0.8\linewidth]{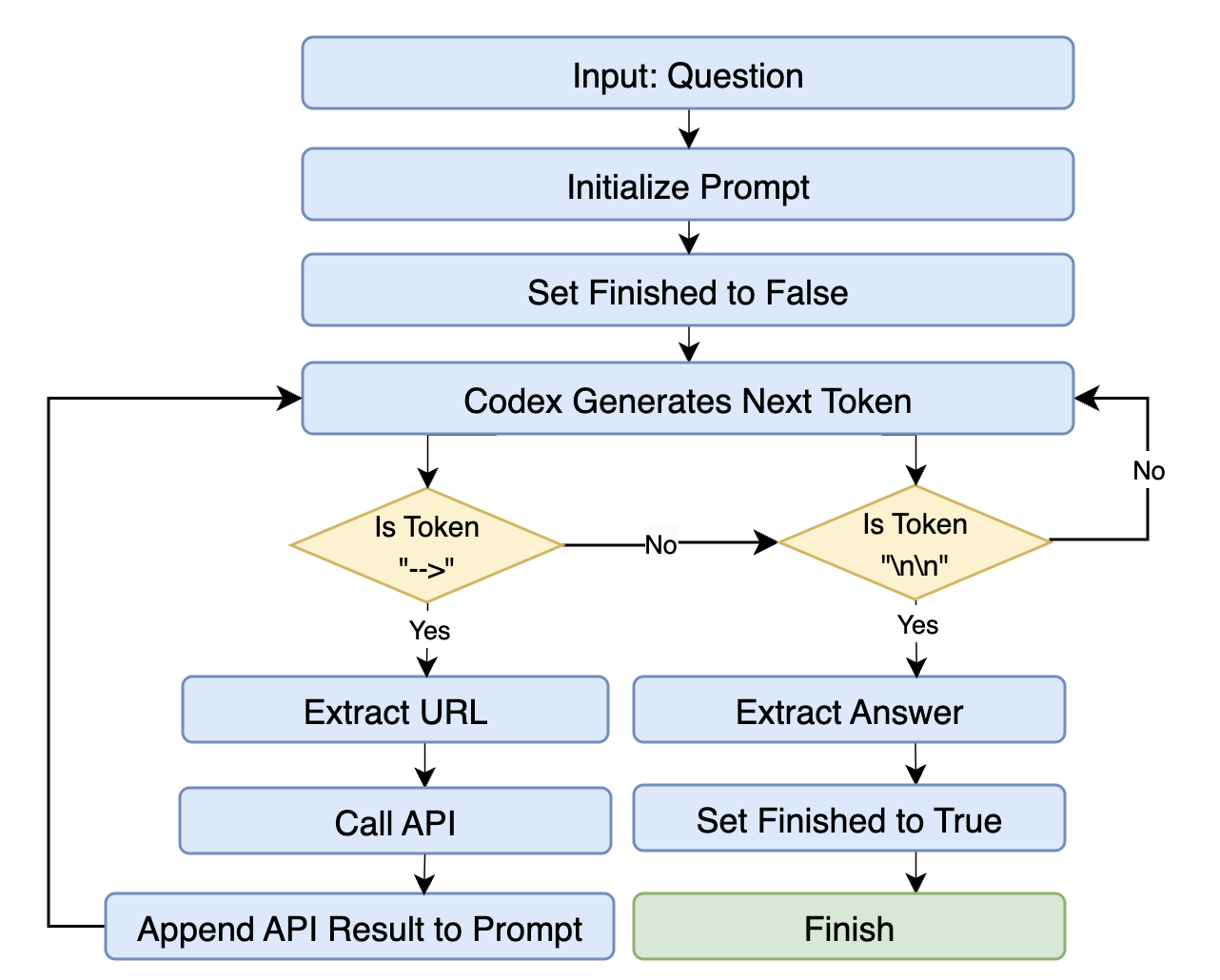}
	\vspace{-0.25cm}
	\caption{Inference algorithm in GeneGPT}
	\label{fig:genegpt_pipline}
	\vspace{-0.4cm}
\end{figure}

GeneGPT's prompt comprises four components: a task description, API documentation, example demonstrations, and a test question. There are two versions of GeneGPT: a \textbf{-full} version, which includes two API documentation blocks and four demonstrations, and a \textbf{-slim} version, which omits documentation and includes only two demonstrations. For GeneHop, the prompt was further tailored per task, depending on whether information retrieval required only E-utils or both E-utils and BLAST (see Table~\ref{tab:genehop_inNout}). These components guide the inference process, as illustrated in Figure~\ref{fig:genegpt_pipline}. Table~\ref{tab:gene_components} summarizes the APIs and example usages embedded in the prompt.

\begin{table}[t!]
	\centering
	\caption{Summary of NCBI API usage documentations and demonstrations in the GeneGPT prompt. }
	\label{tab:gene_components}
	\vspace{-0.25cm}
	\resizebox{\linewidth}{!}{
		\begin{tabular}{ccc|cc|cc}
			\hline
			\multirow{2}{*}{\textbf{Doc/Demo}} & \multirow{2}{*}{\textbf{Database}} & \multirow{2}{*}{\textbf{Function}} & \multicolumn{2}{c|}{\textbf{GeneTuring}} & \multicolumn{2}{c}{\textbf{GeneHop}} \\
			\cline{4-7}
			& & & \textbf{-full} & \textbf{-slim} & \textbf{e-utils} & \textbf{blast + eutils} \\ \hline
			E-utils        & gene, snp, omim & esearch, efetch, esummary  & \checkmark & $\circ$ & \checkmark & \checkmark \\ 
			BLAST          & nt              & blastn                      & \checkmark & $\circ$ & $\circ$ & \checkmark \\ 
			\hdashline
			Alias         & gene           & esearch $\rightarrow$ efetch  & \checkmark & \checkmark & \checkmark & \checkmark \\ 
			Gene SNP      & snp            & esummary                     & \checkmark & $\circ$ & \checkmark & $\circ$ \\ 
			Gene disease  & omim           & esearch $\rightarrow$ esummary & \checkmark & $\circ$ & \checkmark & \checkmark \\ 
			Alignment     & nt             & blastn                        & \checkmark & \checkmark & $\circ$ & \checkmark \\ \hline
		\end{tabular}%
	}
\end{table}

\subsubsection*{The Irreproducibility of GeneGPT}

Although GeneGPT achieved strong performance by prompting Codex to interact with NCBI APIs, it is no longer reproducible. The Codex models used in its pipeline were officially shut down in 2024 and are no longer publicly available.\footnote{\url{https://platform.openai.com/docs/deprecations\#instructgpt-models}} This limitation highlights a broader concern: reliance on proprietary models risks future irreproducibility once those models are deprecated. To address this, we re-implement the GeneGPT pipeline using open-source models, ensuring that the system remains accessible for future research in genomic QA.

\vspace{-0.1cm}
\section{Pilot Study: Reproducing GeneGPT}


\vspace{-0.1cm}
\subsection{Open-source LLMs with GeneGPT}

To assess the feasibility of open-source LLMs for genomic QA, we considered mainstream open-source models with strong multi-hop reasoning, RAG, and tool-use capabilities. Based on these criteria, we selected Llama3.1, Qwen2.5, and Qwen2.5-Coder.

Llama3.1 has been widely applied across various domains, particularly in healthcare and biomedical research. Prior works have shown that Llama3.1-70B improves accuracy in medical QA, pharmacogenomic data extraction, and biomedical research, making it a strong candidate for handling genomic queries~\cite{Yang:2025:LLM-MedQ,Li:2024:Enhancin,Tao:2024:Fine-tun}. Additionally, its effectiveness in multi-hop reasoning tasks~\cite{He:2025:MINTQA:-} aligns well with GeneHop, reinforcing its suitability for complex information retrieval and synthesis.

Qwen2.5 has demonstrated superior multi-hop reasoning and tool-use capabilities compared to Llama3.1, offering greater accuracy and stability with a lower error rate~\cite{Ye:2025:ToolHop:}. It also outperforms other open-source models 
low-resource language understanding~\cite{Kostiuk:2025:The-Veln}. Given that genomic datasets often require retrieval from structured sources, such as NCBI, Qwen2.5's proficiency in RAG makes it particularly well-suited for our study.

Qwen2.5-Coder extends Qwen2.5's capabilities with enhanced code generation and structured data processing, making it more effective for API-based tool use and structured retrieval tasks. Since GeneGPT relies on precise API calls to retrieve genomic information from NCBI, a model optimized for coding tasks and structured retrieval is expected to improve performance. Additionally, prior studies indicate that Qwen2.5-Coder exhibits strong generalization in zero-shot and few-shot settings~\cite{Kostiuk:2025:The-Veln}, which is crucial for handling genomic questions.

\vspace{-0.25cm}
\subsection{Selecting model sizes for evaluation}

To evaluate the impact of model size on achieving GeneGPT-like performance, we conducted an initial test using Llama3.1 with sizes of 8B and 70B. The results indicated that the 8B model was unable to comprehend tasks of this nature, while the 70B model produced performance more aligned with the expected outcomes. Based on these findings, we established 70B as the baseline model size. 

For Qwen2.5, we selected the 72B version to maintain consistency with this baseline for comparison. In contrast, Qwen2.5-Coder, designed specifically for coding tasks, is only available in a 32B configuration. Despite its smaller size, we included Qwen2.5-Coder-32B in the experiment to assess its specialized capabilities and potential applicability to the task. For simplicity, we refer to Llama3.1-70B as Llama, Qwen2.5-72B as Qwen, and Qwen2.5-Coder-32B as Qwen-Coder.

%

\vspace{-0.25cm}
\subsection{Experimenting with GeneGPT prompts}
For GeneTuring, we initially adopted the same standard prompting strategy (\textbf{-full}) and pipeline as GeneGPT, replacing Codex (code-davinci-002) with the three selected open-source LLMs for evaluation. We then considered the best performing open-source LLMs with a compact prompt (\textbf{-slim}), including only two demonstrations (Table~\ref{tab:gene_components}). The experiments were conducted on Google Cloud, utilizing an accelerator-optimized compute engine with an A2 standard instance (a2-highgpu-2g), equipped with 2 NVIDIA A100 GPUs (80 GB GPU memory), 24 vCPUs, and 170 GB instance memory. Model deployment was managed via Ollama~\cite{Chiang:2024:Ollama}, while LlamaIndex~\cite{Liu:2024:LlamaInd} was used to access the Ollama LLM client locally.

\vspace{-0.2cm}

\subsection{Evaluation methods}
To ensure the consistency and reproducibility of the evaluation, we wrote a set of automated scripts to batch analyze and score the model outputs of 9 and 3 tasks respectively for the GeneTuring and GeneHop benchmarks. The overall evaluation criteria remain consistent with the original GeneGPT paper, as shown in Table~\ref{tab:evaluation-criteria}.

However, several tasks involve a certain degree of semantic understanding and subjective judgment, and they cannot be accomplished by simply relying on string matching or recall-based approaches. These include SNP gene function, Multi-species DNA alignment, and Protein-coding genes, for which we introduce an LLM as an evaluator (specifically, the Qwen2.5 32B model) to perform semantic-level scoring\footnote{Implementation details including prompts are available at \url{https://github.com/ielab/OpenBioLLM}}. The model outputs a score of 1, 0.5, or 0 based on the semantic consistency between the candidate answer and the reference answer, which is a more reasonable reflection of the quality of responses in complex tasks. We note that recent work has reported the possibility of LLM-based evaluators to display bias towards content generated by the same LLM backbone used as evaluator~\cite{balog2025rankers}. To ward off the introduction of backbone preferences, we compared the output of our LLM-based evaluator with the labels provided by human assessors for GeneGPT responses and made available by the original authors~\cite{Jin:2024:GeneGPT:}; we find the LLM-based evaluator assessments to match those of the human assessors.


\begin{table}[t]
	\centering
	\vspace{-0.2cm}
	\caption{Evaluation Criteria for GeneTuring and GeneHop}
	\label{tab:evaluation-criteria}
	\vspace{-0.3cm}
	\resizebox{\linewidth}{!}{%
		\begin{tabular}{ll}
			\hline
			\textbf{Task}                          & \textbf{Evaluation Criteria} \\ \hline
			\multicolumn{2}{l}{\textbf{GeneTuring Tasks}} \\
			Gene alias                             & String match \\
			Gene disease association               & Recall (correct answers / ground-truth answers) \\
			Gene location                          & String match \\
			Human genome DNA alignment             & Exact match=1 point; correct chromosome but incorrect position=0.5 point \\
			Multi-species DNA alignment            & String match (species name mapping) \\
			Gene name conversion                   & String match \\
			Protein-coding genes                   & String match (binary mapping of correctness) \\
			Gene SNP association                   & String match \\
			SNP location                           & String match \\ \hline
			\multicolumn{2}{l}{\textbf{GeneHop Tasks}} \\
			SNP gene function                      & Manual semantic judgement (1, 0.5, or 0 point) \\
			Disease gene location                  & Recall (correct answers / ground-truth answers) \\
			Sequence gene alias                    & Recall (correct answers / ground-truth answers) \\ \hline
		\end{tabular}%
	}
	\vspace{-0.3cm}
\end{table}

\vspace{-0.25cm}
\subsection{Effectiveness of GeneGPT with Open LLMs}

\textbf{Sample-Based Evaluation.} Before conducting a full-scale assessment of open LLMs with the GeneGPT pipeline, we performed a pilot evaluation by sampling the first ten questions (10Qs) from the 50 available per task in both GeneTuring and GeneHop. This approach allowed us to assess preliminary model performance before proceeding with optimization and comprehensive evaluation. 

\vspace{0.5em}\noindent\textbf{Results on GeneTuring.} Table~\ref{tab:pilot-results} shows that \textit{Llama-full} significantly underperformed GeneGPT-full, with an average accuracy of only 0.19 versus 0.84. 
In contrast, \textit{Qwen-full 72B} outperformed GeneGPT-full across most tasks, achieving an average accuracy of 0.89. It obtained perfect scores in five tasks and notably outperformed GeneGPT-full in gene-disease association (1.00 vs. 0.85) and protein-coding genes (1.00 vs. 0.60), while performing comparably in gene location and human genome DNA alignment. 

We also evaluated \textit{Qwen-Coder}, a code-focused variant of Qwen2.5, which performed poorly overall. While it could construct API request links, it failed to extract meaningful answers from responses, highlighting a key limitation for code-specialized models in knowledge extraction tasks.

Overall, these pilot evaluations suggest that while Llama struggles with the GeneGPT pipeline, Qwen demonstrates strong potential as a replacement model. These initial results motivated us to use Qwen as the core open LLM backbone for the remainder of the experiments.



\begin{table}[t!]
\centering
\caption{Effectiveness obtained when implementing GeneGPT using three open-LLMs, compared to the original GeneGPT effectiveness using OpenAI's Codex. These results are obtained using only the first ten GeneTuring questions per task.}
\label{tab:pilot-results}
\vspace{-0.3cm}
\resizebox{\linewidth}{!}{%
\begin{tabular}{lcccc}
\hline
\textbf{GeneTuring Task (10Qs)} 
& \textbf{Llama-full} 
& \textbf{Qwen-full} 
& \textbf{Qwen-Coder-full} 
& \textbf{GeneGPT-full} 
\\ \hline
Gene alias & 0.20 & \textbf{1.00} & 0.30 & \textbf{1.00} \\
Gene disease association & 0.10 & \textbf{1.00} & 0.25 & 0.85 \\
Gene location & 0.00 & \textbf{0.70} & 0.00 & \textbf{0.70} \\
Human genome DNA alignment & 0.20 & \textbf{0.50} & \textbf{0.50} & \textbf{0.50} \\
Multi-species DNA alignment & 0.80 & \textbf{0.90} & \textbf{0.90} & \textbf{0.90} \\
Gene name conversion & 0.00 & \textbf{1.00} & 0.10 & \textbf{1.00} \\
Protein-coding genes & 0.10 & \textbf{1.00} & 0.50 & 0.60 \\
Gene SNP association & 0.30 & \textbf{1.00} & 0.20 & \textbf{1.00} \\
SNP location & 0.00 & 0.90 & 0.10 & \textbf{1.00} \\
\hline
\textbf{Average} & 0.19 & \textbf{0.89} & 0.32 & 0.84 \\
\hline
\end{tabular}%
}
\end{table}

\vspace{0.5em} \noindent \textbf{Results on GeneHop}: The pilot evaluation on GeneHop showed that GeneGPT achieved a moderate overall score of 0.62 (see Table~\ref{tab:qwen2.5-genegpt-genehop}). While this aligns with the original GeneGPT study~\cite{Jin:2024:GeneGPT:} (0.50 on full evaluation of 50 questions, outperforming the baseline model New Bing at 0.24), its performance remains suboptimal, leaving room for improvement. 
In contrast, Qwen outperformed GeneGPT across all tasks, underscoring its superior multi-step reasoning capabilities.

\begin{table}[t]
\centering
\vspace{-0.2cm}	
\caption{Effectiveness obtained when implementing GeneGPT using Qwen, compared to the original GeneGPT effectiveness using OpenAI's Codex. These results are obtained using only the first ten GeneTuring questions per task.}
\label{tab:qwen2.5-genegpt-genehop}
\vspace{-0.3cm}	
\resizebox{\linewidth}{!}{%
\begin{tabular}{lccc|ccc}
\hline
\multirow{2}{*}{\textbf{GeneHop Task (10Qs)}} & \multicolumn{3}{c|}{\textbf{Qwen}} & \multicolumn{3}{c}{\textbf{Accuracy}} \\ \cline{2-7} 
                                       & \textbf{Correct} & \textbf{Half} & \textbf{Error} & \textbf{Qwen} & \textbf{GeneGPT}  \\ \hline
SNP gene function                      & 10               & 0             & 0              & \textbf{1.00}             & 0.85                         \\
Disease gene location                  & 7                & 2             & 1              & \textbf{0.80}             & 0.70                          \\
Sequence gene alias                    & 5                & 3             & 2              & \textbf{0.67}             & 0.30                         \\ \hline
\textbf{Average}                       & -       & -    & -     & \textbf{0.82}    & 0.62       \\ \hline
\end{tabular}%
}
\vspace{-0.2cm}	
\end{table}
\vspace{-0.2cm}
\section{Failure patterns and opportunities}
In our pilot experiments involving the reproduction of GeneGPT, Qwen demonstrated great potential in handling tasks similar to those explored in the original GeneGPT study. Building on this, we undertook a detailed error analysis of the reproduction experiments. By manually examining model outputs and investigating the sources of incorrect answers, we identified several recurring failure patterns that point to specific opportunities for refinement.


%

\vspace{-0.1cm}
\subsection{Obs. 1: Exact match \& case sensitivity}

Some errors arise from the benchmark’s strict requirement for exact matching of gene symbols, which conflicts with Qwen’s behavior under the original GeneGPT prompt that handles gene symbols in a case-insensitive manner. For example, in response to the question ``\textit{What is the official gene symbol of GalNAc-T4?}'', Qwen returned ``GALNT4'', which is an alias corresponding to the case-insensitive match ``GALNAC-T4''. However, the ground-truth answer is ``POC1B-GALNT4'', which is the official gene symbol linked to the alias ``GalNAc-T4'' with correct casing. Because the model ignored case distinctions, it selected an alias with a similar surface form but failed to resolve it to the exact official symbol required by the benchmark.




\vspace{-0.1cm}
\subsection{Obs. 2: Relevant answer ranked too low}
Another common issue occurs because the relevant answer appears too far down in the response list. Since the HTTP requests to the NCBI API retrieve only the top 5 or 10 items, it may be excluded entirely. These errors stem from limitations in both the model and the NCBI database.


A concrete example is the question: ``\textit{Which chromosome is the RGS16 gene located on in the human genome?}''. In answering this question, Qwen omitted the ``sort=relevance'' parameter when assembling the HTTP request to the NCBI API. This omission has little effect on small datasets but causes irrelevant results to be ranked higher in large datasets. Unlike Qwen-Coder, the Qwen 72B model rarely overlooks this parameter. Emphasizing its inclusion in the API requests is crucial to mitigating this issue.


%
 
 \vspace{-0.1cm}
 \subsection{Obs. 3: Query difficulty}
Some mistakes can be attributed to ambiguous query interpretation and inconsistent API result ordering. For example, in response to ``\textit{What is the official gene symbol of TFA?}’’, Qwen generated a correct request to the NCBI API, but the response was challenging. TFA is both an official symbol and an alias. Here, the question treats it as an alias, but the API prioritizes the official symbol, pushing the alias reference lower in the list, leading the model to select the wrong answer.

\vspace{-0.1cm}
\subsection{Obs. 4: Discrepancies in coordinate mapping between genome assemblies}
In the DNA alignment task, while Qwen consistently identifies the correct chromosome, the inferred genomic coordinates often deviate from the reference standard. This discrepancy arises because the GeneTuring benchmark defines ground-truth coordinates using the reference assembly of \texttt{BSgenome.Hsapiens.UCSC.hg38}, whereas queries to NCBI's \texttt{core\_nt} database frequently return results based on other assemblies (e.g., GRCh37 or clone-based sequences).  As a result, while chromosome predictions are accurate, their exact positions may not align with the hg38 standard. We note this problem is not specific to Qwen and it affects any model in the GeneGPT pipeline.



\vspace{-0.1cm}
\section{Prompt Engineering for Qwen}
\vspace{-0.05cm}


We now focus on optimizing the Qwen backbone by refining prompts. Overall, we do the following prompt improvements:

\vspace{0.2cm}
\noindent\textbf{1. Emphasize the output format:}  
To improve response consistency, we explicitly instructed the model to follow the standardized output format in Figure~\ref{fig_prompt}. Enforcing this structure ensured accurate, consistent responses while maintaining adherence to the expected format.

\begin{figure}[t]
    \centering
    \begin{tcolorbox}[colback=gray!10, colframe=gray!80, boxrule=0.01pt, arc=4pt, width=\columnwidth]
	\begin{enumerate}[labelindent=0pt, leftmargin=1em, noitemsep, topsep=0pt]
		\item[1.] When you need to make an API call, please put the URL you generated directly in \texttt{[]}. Do not add anything else.
		\item[2.] When you reach the final answer, please use the following format: \texttt{Answer: your generated answer.}
	\end{enumerate}
\end{tcolorbox}
\vspace{-14pt}
    \caption{Snippet of the prompt component used in OpenBioLLM to enforce standardized output format.}
    \vspace{-10pt}
    \label{fig_prompt}
\end{figure}

\vspace{0.1cm}
\noindent\textbf{2. Correct the missing parameters:} 
A significant portion of errors stemmed from queries that failed to retrieve the most relevant results. This was due in part to an inconsistency in the original prompt design: while the demonstration examples included the ``sort=relevance'' parameter, the accompanying documentation omitted it. This mismatch created ambiguity, especially for smaller models (e.g., 14B and 32B), making it unclear when the parameter should be used. To address this, we updated the documentation to explicitly include ``sort=relevance'' for consistency and reliability.

Additionally, we added the missing ``retmode=json'' parameter to the documentation. By default, the NCBI API returns responses in XML format, which poses two main issues: JSON is easier for LLMs to parse accurately, and XML responses are longer, increasing token usage. For instance, a gene esummary API query returns 5,659 tokens in JSON but 6,820 in XML. Switching to JSON reduces token consumption, lowers processing overhead, and allows more information to fit within the model’s context window.

\vspace{0.1cm}
\noindent\textbf{3. Incorporating ReAct-style Prompting:}
To better tackle multi-hop reasoning tasks in GeneHop, we draw inspiration from the ReAct prompting framework~\cite{Yao:2023:ReAct:-S}, which interleaves \textit{reasoning} steps with \textit{actions}. In ReAct, LLMs are prompted to think step-by-step (“Thought”), execute external tool calls (“Action”), observe results (“Observation”), and make iterative decisions accordingly. This approach improves transparency, control, and performance in complex workflows.

Building on this principle, we design domain-specific ReAct-style prompts to explicitly guide the model through sub-question decomposition and tool interaction. Unlike vanilla prompting, our design not only teaches the model how to decompose complex biomedical queries, but also explicitly instructs it on what to do after receiving a tool response—e.g., which field to extract or what the next reasoning step should be. 

\subsection{Evaluation of Prompt Optimization}
\subsubsection*{GeneTuring} After refining the prompt, we evaluated it on the GeneTuring benchmark by running all nine tasks (450 questions in total) with Qwen-32B and Qwen-72B using the full set of API documentations and demonstrations. Both models achieved a macro-average accuracy of 0.838, which was the highest among the settings tested at this stage. These results demonstrate the effectiveness of our optimized prompt design. Detailed results are presented in Table~\ref{tab:qwen2.5-32b-gene-optimized}.

\begin{table}[t]
\renewcommand{\arraystretch}{1} 
\setlength{\tabcolsep}{3pt}       
    \centering
    \vspace{-0.1cm}
    \caption{Effectiveness obtained on GeneTuring after Qwen models were prompt optimizated. GeneGPT's results are from \cite{Jin:2024:GeneGPT:} and were obtained with the original prompt.}
    \label{tab:qwen2.5-32b-gene-optimized}
    \vspace{-0.2cm}
    \resizebox{\linewidth}{!}{ 
	\begin{tabular}{lcccccc|cccc}
        \hline
        \multirow{3}{*}{\textbf{GeneTuring Task}}  
            & \multicolumn{6}{c|}{\textbf{Qwen-full}} 
            & \multicolumn{4}{c}{\textbf{Accuracy}} \\ 
        \cline{2-11}
            & \multicolumn{2}{c}{\textbf{Correct}} 
            & \multicolumn{2}{c}{\textbf{Half}} 
            & \multicolumn{2}{c|}{\textbf{Errors}}  
            & \multicolumn{2}{c}{\textbf{Qwen-full}}  
            & \multicolumn{2}{c}{\textbf{GeneGPT} } \\ 
        \cline{2-11}
            & \textbf{\footnotesize{32B}} & \textbf{\footnotesize{72B}} 
            & \textbf{\footnotesize{32B}} & \textbf{\footnotesize{72B}} 
            & \textbf{\footnotesize{32B}} & \textbf{\footnotesize{72B}} 
            & \textbf{\footnotesize{32B}} & \textbf{\footnotesize{72B}} 
            & \textbf{\footnotesize{full}} & \textbf{\footnotesize{slim}} \\ 
        \hline
        Gene alias                   & 45  & 44  & 0  & 0  & 5  & 6  & \textbf{0.90}  & 0.88  & 0.80 & 0.84 \\ 
        Gene disease association     & 34  & 34  & 5  & 3  & 11 & 13  & 0.73  & 0.71  	& \textbf{0.76} & 0.70  \\
        Gene location                & 35  & 36  & 0  & 0 & 15 & 14  & 0.70  & \textbf{0.72}  & 0.62 & 0.66  \\ 
        Human genome DNA alignment   & 0   & 0   & 45 & 45 & 5  & 5  & \textbf{0.45}  & \textbf{0.45}  & 0.44 & 0.44 \\ 
        Multi-species DNA alignment  & 41  & 40  & 0  & 0  & 9  & 10  & 0.82  & 0.80  & 0.88 & \textbf{0.90}  \\ 
        Gene name conversion         & 50  & 50  & 0  & 0  & 0  & 0  & \textbf{1.00}  & \textbf{1.00}  	& \textbf{1.00} & \textbf{1.00} \\
        Protein-coding genes         & 50  & 49  & 0  & 0  & 0  & 1  & \textbf{1.00}  & 0.98  & 0.74 & 0.98  \\ 
        Gene SNP association         & 50  & 50  & 0  & 0  & 0  & 0  & \textbf{1.00}  & \textbf{1.00}  	& \textbf{1.00} & \textbf{1.00} \\
        SNP location                 & 47  & 50  & 0  & 0  & 3  & 0  & 0.94  & \textbf{1.00}  & \textbf{1.00} & 0.98   \\
        \hline
        \textbf{Average}             & -   & -   & -  & -  & -  & -  & \textbf{0.838}  & \textbf{0.838}  & 0.804 & 0.833    \\ 
        \hline
    \end{tabular}
	}
    \vspace{-0.3cm}
\end{table}


\subsubsection*{GeneHop}
Table~\ref{tab:qwen2.5-32b-hop-optimized} summarizes the optimized performance of Qwen models on GeneHop. Substantial improvements were observed over GeneGPT, particularly in the sequence gene alias task, where scores rose from 0.35 (GeneGPT) to 0.75 (32B) and 0.71 (72B). In the SNP gene function task, the 72B model achieved the highest score (0.93), outperforming both its smaller counterpart and GeneGPT. However, performance on the disease gene location task was more mixed: the 32B model scored only 0.17, while the 72B model reached 0.54—still below GeneGPT’s 0.67. On average, the 72B variant led across tasks, highlighting the combined benefits of prompt optimization and increased model size for GeneHop.
\begin{table}[t]
\renewcommand{\arraystretch}{1} 
\setlength{\tabcolsep}{3pt}       
	\centering
	\caption{Effectiveness obtained on GeneHop using the ``ReAct-style'' prompt. GeneGPT's results are from \cite{Jin:2024:GeneGPT:} and were obtained with the original prompt.}
	\label{tab:qwen2.5-32b-hop-optimized}
	\vspace{-0.2cm}
	\resizebox{\linewidth}{!}{%
       \begin{tabular}{ccccccc|cccc}
            \hline
            \multirow{3}{*}{\textbf{GeneHop Task}}  
                & \multicolumn{6}{c|}{\textbf{Qwen}} 
                & \multicolumn{3}{c}{\textbf{Accuracy}} \\ 
            \cline{2-11}
                & \multicolumn{2}{c}{\textbf{Correct}} 
                & \multicolumn{2}{c}{\textbf{Half}} 
                & \multicolumn{2}{c|}{\textbf{Error}} 
                & \multicolumn{2}{c}{\textbf{Qwen}}  
                & \multirow{2}{*}{\textbf{GeneGPT}} \\ 
            \cline{2-9}
                & \textbf{\footnotesize{32B}} &\textbf{\footnotesize{72B}} 
                & \textbf{\footnotesize{32B}} & \textbf{\footnotesize{72B}} 
                & \textbf{\footnotesize{32B}} & \textbf{\footnotesize{72B}} 
                & \textbf{\footnotesize{32B}} &\textbf{\footnotesize{72B}} &  \\ 
            \hline

            SNP gene function         & 40  & 45  & 7  & 3  & 3  & 2  & 0.87  & \textbf{0.93}  & 0.61 \\ 
            Disease gene location     & 8  & 24  & 1   & 6  & 41  & 20 & 0.17  & 0.54  & \textbf{0.67} \\ 
            Sequence gene alias       & 34  & 16  & 8   & 27  & 8  & 7  & \textbf{0.75}  & 0.71  & 0.35 \\ 
            \hline

            \textbf{Average}                   & -   & -   & -   & -  & -   & -  & 0.597  & \textbf{0.729}  & 0.543 \\ 
            \hline
        \end{tabular}%
	}
	\vspace{-10pt}
\end{table}

\vspace{-6pt}
\subsection{Ablation Study}
GeneGPT’s prompt structure is summarized in Table~\ref{tab:gene_components}. In the original ablation study by Jin et al.~\cite{Jin:2024:GeneGPT:}, GeneGPT demonstrated that a streamlined prompt containing only two demonstrations (referred to as the slim version) generalized well across tasks, sometimes even outperforming the full version.

We adopted a similar slim prompt setup for Qwen, including only two demonstrations: gene alias resolution and DNA alignment. While Qwen-72B-slim achieved performance comparable to its full version, Qwen-32B-slim saw a significant drop, from 0.838 to 0.467, see Table~\ref{tab:qwen2.5-32b-gene}. This suggests that larger models exhibit stronger cross-task generalization under prompt compression.

\begin{table}[t]
\renewcommand{\arraystretch}{1} 
\setlength{\tabcolsep}{3pt}       
\centering
\caption{Ablation results on GeneTuring.}
\label{tab:qwen2.5-32b-gene}
\vspace{-0.4cm}
\resizebox{\linewidth}{!}{%
\begin{tabular}{lcccccc|cccccc}
\hline
\multirow{3}{*}{\textbf{GeneTuring Task}} & \multicolumn{6}{c|}{\textbf{Qwen-slim}} & \multicolumn{6}{c}{\textbf{Accuracy}} \\ \cline{2-13}
& \multicolumn{2}{c}{\textbf{Correct}} & \multicolumn{2}{c}{\textbf{Half}} & \multicolumn{2}{c|}{\textbf{Errors}} & \multicolumn{4}{c}{\textbf{Qwen}} & \multicolumn{2}{c}{\textbf{GeneGPT$^\dagger$}} \\ \cline{2-13}
& \textbf{32B} & \textbf{72B} & \textbf{32B} & \textbf{72B} & \textbf{32B} & \textbf{72B} & \textbf{32B full} & \textbf{32B slim} & \textbf{72B full} & \textbf{72B slim} & \textbf{full} & \textbf{slim} \\ \hline
Gene alias                        & 37 & 43 & 0 & 0 & 13 & 7 & \textbf{0.90} & 0.74   & 0.88 & 0.86 & 0.80 & 0.84 \\
Gene disease association          & 31 & 36 & 6 & 2 & 13 & 12 & 0.73 & 0.69  & 0.71 & 0.75  & \textbf{0.76} & 0.70  \\
Gene location                     & 7 & 28 & 0 & 0 & 43 & 22 & 0.70 & 0.14  & \textbf{0.72} & 0.56  & 0.62 & 0.66  \\
Human genome DNA align.           & 0 & 0 & 39 & 47 & 11 & 3 & 0.45 & 0.39  & 0.45 & \textbf{0.47}  & 0.44 & 0.44  \\
Multi-species DNA align.          & 40 & 38 & 0 & 0 & 10 & 12 & 0.82 & 0.80  & 0.80 & 0.76  & 0.88 & \textbf{0.90}  \\
Gene name conversion              & 0 & 49 & 0 & 0 & 50 & 1 & \textbf{1.00} & 0.00  & \textbf{1.00} & 0.98  & \textbf{1.00} & \textbf{1.00} \\
Protein-coding genes              & 45 & 48 & 0 & 0 & 5 & 2 & \textbf{1.00} & 0.90  & 0.98 & 0.96  & 0.74 & 0.98  \\
Gene SNP association              & 16 & 50 & 0 & 0 & 34 & 0 & \textbf{1.00} & 0.32  & \textbf{1.00} & \textbf{1.00} & \textbf{1.00} & \textbf{1.00} \\
SNP location                      & 11 & 49 & 0 & 0 & 39 & 1 & 0.94 & 0.22  & \textbf{1.00} & 0.98  & \textbf{1.00} & 0.98  \\ \hline
\textbf{Average}                           & - & - & - & - & - & - & \textbf{0.838} & 0.467  & \textbf{0.838} & 0.813  & 0.804 & 0.833  \\ \hline
\end{tabular}
}
\vspace{-0.5cm}
\end{table} 

\section{Multi-Agent Architecture}
\subsection{Why a multi-agent architecture?}
Although prompt engineering improves accuracy, the single-model architecture still faces fundamental limitations.

First, multi-step tasks require inserting all API results, including large JSON outputs, directly into the prompt. This greatly increases the context size, making it harder for the model to focus on the most relevant information, even with a large context window.

Second, assigning all responsibilities—from building API calls to extracting information and generating answers—to one model might not be effective. Different tasks benefit from different capabilities.
Dividing responsibilities across specialized modules would likely improve effectiveness.

Finally, single-model systems are difficult to interpret and debug. When the answer is wrong, it is unclear whether the issue lies in the API call, data retrieval, or final reasoning step. This lack of transparency limits our ability to diagnose and improve the system.

\vspace{-0.2cm}
\subsection{Modular multi-agent pipeline}
To address the limitations of a monolithic design, we propose a modular multi-agent architecture, \textit{OpenBioLLM}, for genomic QA. Rather than relying on a single large model to handle all stages of reasoning, our approach decomposes the workflow into specialized agents, each responsible for a distinct task. This modularization enables more efficient use of model capabilities and improves both interpretability and traceability by exposing intermediate reasoning steps.

The pipeline is implemented using the LangGraph framework and comprises six core components: three pipeline controllers—\textit{Router}, \textit{Evaluator}, and \textit{Generator}—and three tool agents—\textit{Eutils Agent}, \textit{BLAST Agent}, and \textit{Web Search Agent}.

The process begins with a user query, which is routed by the \textit{Router} to the appropriate Tool Agent based on the query type. The selected Tool Agent retrieves relevant information and sends it to the \textit{Evaluator}, which determines whether the information is sufficient to answer the question. If not, the query is returned to the Router for further processing. Once sufficient evidence has been gathered, the \textit{Generator} synthesizes the final answer.

\begin{figure}[h!]
    \vspace{-0.1cm}
	\centering
	\includegraphics[width=0.8\linewidth]{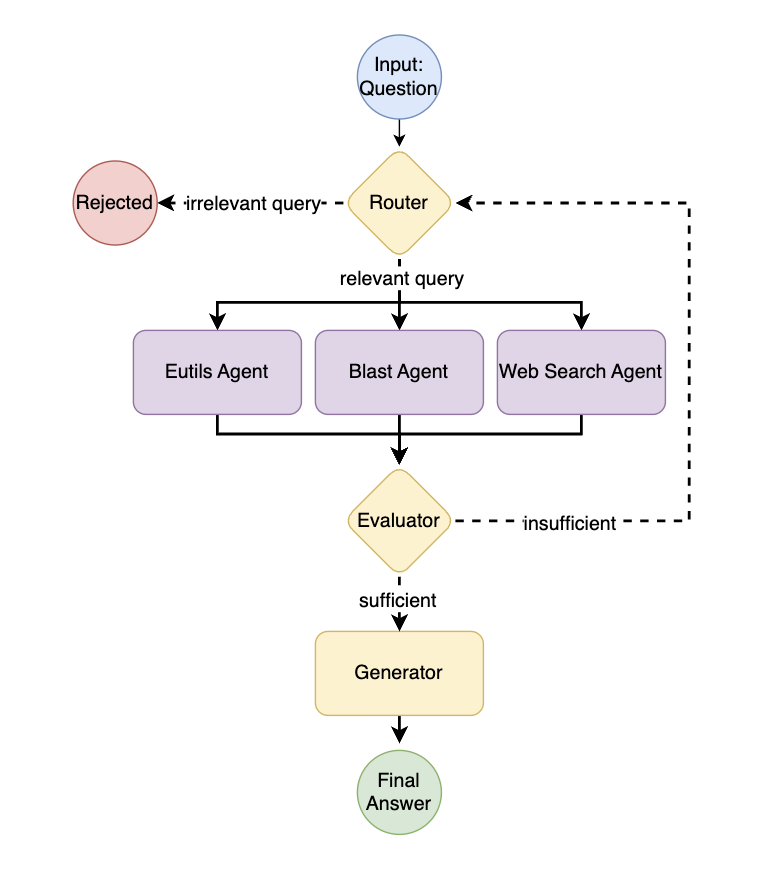}
    \vspace{-0.35cm}
	\caption{Multi-Agent Pipeline in OpenBioLLM Framework.}
	\label{fig:multi_agent_pipeline}
    \vspace{-14pt}
\end{figure}

An overview of the architecture is provided in Figure~\ref{fig:multi_agent_pipeline}. Next, we detail the roles of each component, starting with the three pipeline controllers, followed by the three tool agents.

\vspace{0.5em}\noindent\textbf{Router Module:}
The Router determines which Agent to invoke based on the user question, dialog history, and Evaluator feedback. It routes requests to one of three Agents: Eutils, Blast, or Web Search. If a question is unrelated to bioinformatics, the Router generates a polite rejection to maintain system professionalism and stability.



\vspace{0.5em}\noindent\textbf{Evaluator Module:}
The Evaluator makes decisions based on the user question, conversation history, and the outputs of invoked API tools. To improve interpretability and stability, its output follows a strict JSON format with two fields: \texttt{next\_step} (the decided action) and \texttt{reason} (the explanation). This explicit reasoning enhances transparency, traceability, and encourages the model to reflect before proceeding.



\vspace{0.5em}\noindent\textbf{Generator Module:}
In the Generator module, the system generates the final answer by combining the user query with results from Eutils, BLAST, and Web Search. To ensure professionalism and clarity, the model is prompted to act as a bioinformatician, providing clear and concise answers that cite specific facts (e.g., gene names, sequences) and acknowledge any limitations when information is incomplete.


\vspace{0.5em}\noindent\textbf{Eutils Agent Module:}
In the Eutils Agent module, the system retrieves information through a two-step process: it first obtains relevant IDs and then fetches detailed records based on those IDs. Unlike the original GeneGPT design, which required generating full HTTP URLs, our approach outputs structured JSON parameter objects. This reduces token usage, improves traceability by recording parameter combinations, and avoids redundant requests. Hard-coded defaults (e.g., \texttt{retmode=json}, \texttt{sort=relevance}) ensure essential arguments are always included. The agent also skips requests with previously used parameters to enhance efficiency.

\vspace{0.5em}\noindent\textbf{BLAST Agent Module:}
The BLAST Agent compares user-supplied DNA sequences with the reference genome through a two-step process: first, it submits the sequence via a PUT request to obtain a request ID (RID); then, after about 10 seconds, it polls the result with a GET request using the RID.
To improve robustness and efficiency, the agent records processed sequences to avoid duplicate submissions and includes wait-and-retry logic, polling every 10 seconds and retrying up to three times if results are not ready.
In the prompt design, the model is guided to extract and validate DNA sequences from user input, outputting them in strict JSON format. Only outputs containing a valid “sequence” field are accepted.



\vspace{0.5em}\noindent\textbf{Web Search Agent Module:}
The Web Search Agent uses the Google Custom Search API to retrieve real-time information for queries that cannot be answered by structured databases like Eutils or BLAST. It serves as a fallback for out-of-domain questions, ambiguous inputs, and unstructured queries such as general biological facts or conceptual topics.
Unlike other agents, the Web Search Agent operates as a purely functional component: it formulates a search request based on the user question and retrieves the top 5 results, extracting key fields such as title, source, date, author, summary, and keywords.



\vspace{-0.2cm}
\section{Evaluation of OpenBioLLM}
\begin{table*}[t]
\renewcommand{\arraystretch}{1} 
\setlength{\tabcolsep}{3pt}       
\centering
\caption{Effectiveness of OpenBioLLM on GeneTuring and GeneHop across Qwen2.5 models of varying size. Comparisons with GeneGPT are drawn from~\cite{Jin:2024:GeneGPT:} for the OpenAI-based model, and from Section 6 for the Qwen-based model. For instances denoted with Multi-, the first value indicates the Pipeline Controller’s model size, while the second value indicates the Tool Agents’ model size.(e.g., ‘14b+7b’:  14B Pipeline Controller and 7B Tool Agents).}
\label{tab:performance_comparison}
\vspace{-0.25cm}
\resizebox{\textwidth}{!}{%
\begin{tabular}{l|cc|cc:cc|cccc}
\hline
\multirow{2}{*}{\textbf{Model}}
 & \multicolumn{2}{c|}{\textbf{Original-GeneGPT (code-davinci-002)}}& \multicolumn{4}{c|}{\textbf{Optmized-GeneGPT (Qwen2.5)}} & \multicolumn{4}{c}{\textbf{OpenBioLLM (Qwen2.5)}} \\
\cline{2-11}
 & \textbf{GeneGPT-full} & \textbf{GeneGPT-slim} & \textbf{Qwen32b-full} & \textbf{Qwen32b-slim} & \textbf{Qwen72b-full} & \textbf{Qwen72b-slim} & \textbf{Multi-(14b+7b)} & \textbf{Multi-(14b+14b)} & \textbf{Multi-(32b+14b)} & \textbf{Multi-(32b+32b)} \\
\hline
\textbf{GeneTuring Tasks}& & & & & & & & & &\\
Gene alias & 0.80 & 0.84 & 0.90 & 0.74 & 0.88 & 0.86 & 0.84 & 0.90 & 0.92 & \textbf{0.96} \\
Gene disease association & 0.76 & 0.70 & 0.73 & 0.69 & 0.71 & 0.75 & 0.48 & \textbf{0.81} & 0.76 & 0.73 \\
Gene location & 0.62 & 0.66 & 0.70 & 0.14 & \textbf{0.72} & 0.56 & 0.68 & 0.66 & 0.66 & 0.66 \\
Human genome DNA alignment & 0.44 & 0.44 & 0.45 & 0.09 & 0.45 & 0.47 & 0.47 & 0.47 & 0.46 & \textbf{0.48} \\
Multi-species DNA alignment & 0.88 & \textbf{0.90} & 0.82 & 0.80 & 0.80 & 0.76 & 0.84 & 0.84 & 0.82 & 0.82 \\
Gene name conversion & \textbf{1.00} & \textbf{1.00} & \textbf{1.00} & \textbf{1.00} & \textbf{1.00} & 0.98 & \textbf{1.00} & \textbf{1.00} & \textbf{1.00} & \textbf{1.00} \\
Protein-coding genes & 0.74 & 0.71 & 0.90 & 0.90 & 0.98 & 0.96 & \textbf{1.00} & 0.98 & \textbf{1.00} & \textbf{1.00} \\
Gene SNP association & \textbf{1.00} & \textbf{1.00} & \textbf{1.00} & 0.32 & \textbf{1.00} & \textbf{1.00} & 0.94 & \textbf{1.00} & \textbf{1.00} & \textbf{1.00} \\
SNP location & \textbf{1.00} & 0.94 & 0.22 & \textbf{1.00} & 0.98 & 0.98 & 0.98 & 0.98 & 0.98 & 0.98 \\
\textbf{Average} & 0.804 & 0.833 & 0.838 & 0.467 & 0.838 & 0.813 & 0.799 & \textbf{0.849} & 0.844 & 0.848 \\
\hline
\textbf{GeneHop Tasks}& & & & & & & & & &\\
SNP gene function
& \multicolumn{2}{c|}{0.61}
& \multicolumn{2}{c:}{0.87}
& \multicolumn{2}{c|}{\textbf{0.93}}
& 0.63 & 0.82 & 0.85 & 0.78 \\

Disease gene location
& \multicolumn{2}{c|}{0.67}
& \multicolumn{2}{c:}{0.17}
& \multicolumn{2}{c|}{0.54}
& 0.22 & 0.75 & \textbf{0.79} & 0.69 \\
Sequence gene alias
& \multicolumn{2}{c|}{0.35}
& \multicolumn{2}{c:}{0.75}
& \multicolumn{2}{c|}{0.85}
& 0.72 & 0.72 & 0.85 & \textbf{0.91} \\
\textbf{Average}
& \multicolumn{2}{c|}{0.543}
& \multicolumn{2}{c:}{0.597}
& \multicolumn{2}{c|}{0.727}
& 0.431 & 0.761 & \textbf{0.830} & 0.793 \\
\hline
\end{tabular}
}
\end{table*}

To our multi-agent OpenBioLLM architecture, we tested different model sizes and configurations across system modules. Both the Pipeline Controller components (Router, Evaluator, Generator) and Tool Agents (Eutils, BLAST, Web Search) rely on LLMs, but their roles differ: Tool Agents mainly extract parameters and call APIs, while the Pipeline Controller manages system logic, tool selection, and answer generation. 
We explored the trade-off between performance and computational cost with four configurations, leveraging the availability of Qwen2.5 backbones of different sizes: (1) a lightweight setup using 14B for the Controller and 7B for Agents, (2) a balanced setup with 14B for both, (3) an enhanced setup with 32B for the Controller and 14B for Agents, and (4) a fully scaled setup using 32B models throughout. 

Evaluation results are summarized in Table~\ref{tab:performance_comparison}. The results reveal several insightful trends, described below.


\vspace{0.5em}\noindent\textbf{1. Smaller models can achieve competitive performance:}
Surprisingly, using 14B models for both the Pipeline Controller and Tool Agents already yields strong results, slightly outperforming the original GeneGPT’s single 72B model (average GeneTuring score: 0.849 vs. 0.838). This highlights the advantage of modular decomposition: assigning specialized tasks to moderate-sized models can surpass a monolithic LLM with far more parameters.


\vspace{0.5em}\noindent\textbf{2. Larger controllers greatly improve GeneHop performance:}
Using a 32B model for the Pipeline Controller while keeping the Tool Agents at 14B leads to substantial gains on the more complex GeneHop benchmark, raising the average accuracy to 0.830. We attribute this improvement to the controller’s enhanced ability to handle multi-step reasoning, assess sufficiency, and manage fallback logic. As GeneHop tasks demand deeper inference and multi-hop coordination, a stronger controller better orchestrates the multi-agent workflow.


\vspace{0.5em}\noindent\textbf{3. Larger Tool Agents do not yield additional benefits:}
Unexpectedly, upgrading both the controller and Tool Agents to 32B models does not improve performance. GeneTuring results plateau at 0.848, while GeneHop scores slightly decrease. This suggests diminishing returns from scaling Tool Agents, whose tasks are relatively simple and do not require extensive reasoning. It also raises potential issues of redundant computation and increased latency due to unnecessarily large submodules.


\vspace{0.5em}\noindent\textbf{4. Optimal trade-off -- 32B Controller with 14B Tools:}
Overall, the best balance between performance and efficiency is achieved with a 32B model for the Pipeline Controller and 14B models for the Tool Agents. This setup delivers the highest average performance across both GeneTuring and GeneHop, while keeping resource usage reasonable (latency is investigated in Section~\ref{sec_latency}). It underscores the importance of allocating model capacity based on task complexity: using larger models for reasoning and coordination, and smaller models for structured extraction or simple lookups.


\subsection{Error Analysis (32B+14B Multi-Agent)}
We manually analyzed errors from the GeneTuring and GeneHop benchmarks under the 32B+14B configuration. Table~\ref{tab:error_analysis} summarizes the results. Errors were grouped into six categories (E1–E5 and “Other”), each reflecting a specific issue in reasoning or tool usage. Below, we describe each error type and note where it most commonly occurred.

\begin{table}[h!]
\centering
\caption{Error analysis on Multi-Agent Qwen2.5 (32B + 14B)}
\label{tab:error_analysis}
\vspace{-0.25cm}
\small
\begin{tabular}{lrrrrrr}
\toprule
\textbf{GeneTuring Task} & \textbf{E1} & \textbf{E2} & \textbf{E3} & \textbf{E4} & \textbf{E5} & \textbf{O} \\
\midrule
Gene alias                              & 0 & 0 & 2 & 2 & 0 & 0 \\
Gene disease association                & 0 & 0 & 0 & 7 & 0 & 8 \\
Gene location                           & 0 & 0 & 0 & 17 & 0 & 0 \\
Human genome DNA alignment              & 0 & 0 & 0 & 49 & 0 & 1 \\
Multi-species DNA alignment             & 0 & 0 & 0 & 5 & 0 & 4 \\
Gene name conversion                    & 0 & 0 & 0 & 0 & 0 & 0 \\
Protein-coding genes                    & 0 & 0 & 0 & 0 & 0 & 0 \\
Gene SNP association                    & 0 & 0 & 0 & 0 & 0 & 0 \\
SNP location                            & 0 & 0 & 0 & 1 & 0 & 0 \\
\midrule
\textbf{GeneHop Task} & \textbf{E1} & \textbf{E2} & \textbf{E3} & \textbf{E4} & \textbf{E5} & \textbf{O} \\
\midrule
SNP gene function                       & 0 & 15 & 0 & 0 & 0 & 0 \\
Disease gene location                   & 0 & 7  & 0 & 5 & 0 & 1 \\
Sequence gene alias                     & 1 & 4 & 0 & 3 & 2 & 0 \\
\bottomrule
\end{tabular}

\vspace{0.5em}
\footnotesize
\begin{tabular}{@{}l@{}}
E1: Wrong API \quad
E2: Wrong arguments \quad
E3: Wrong comprehension \\
E4: Unanswerable with API \quad
E5: Insufficient query \quad
O: Others
\end{tabular}
\vspace{-0.4cm}
\end{table}

\noindent\textbf{E1: Incorrect API Selection:} These errors occur when the model selects the wrong API for the given task. For example, it may call the BLAST interface when E-utilities is required, or vice versa.


\vspace{0.5em}\noindent\textbf{E2: Incorrect Parameter Usage:}
The model selects the correct API but fails due to incorrect parameters—for example, using the wrong search term, confusing databases like \texttt{db=gene} and \texttt{db=omim}, or misusing \texttt{efetch} and \texttt{esummary}. This error is especially common in GeneHop, with 15 cases in the SNP gene function task and several in Disease gene location.


\vspace{0.5em}\noindent\textbf{E3: Misunderstanding Task Semantics:}
These errors occur when the model misinterprets the question's intent despite having access to the correct information. For example, in the Gene alias task, it may treat the alias as the official name and return it directly, leading to an incorrect answer.


\vspace{0.5em}\noindent\textbf{E4: Insufficient Information to Answer:}
This is the most common error type across both benchmarks. It occurs when the model follows the correct procedure and uses the right tools, but the retrieved data lacks sufficient detail to produce a valid answer. This issue is particularly prevalent in tasks like Human genome DNA alignment and Gene location in GeneTuring, and Disease gene location in GeneHop.

Notably, adding a Web Search Agent only partially alleviates this problem. For example, for sequence-based questions such as ``\textit{Which organism does the DNA sequence come from: AAAAATAATTTCCCGTTAACTGTTAATAAGTATTAGCAG…}'', even searching the full sequence on Google yields little useful information. This highlights a limitation in the biomedical domain: some queries are simply not ``searchable'', rendering external retrieval ineffective in these cases.


\vspace{0.5em}\noindent\textbf{E5: Incomplete Retrieval or Premature Termination:}
These errors occur when the model ends its reasoning too early, leading to incomplete answers. For example, in the Sequence gene alias task, it may retrieve some aliases in the first step but fail to issue additional queries for the full list, resulting in partial results.


\vspace{0.5em}\noindent\textbf{O: Other Errors (Scoring and Evaluation Issues):}
This category includes errors arising from evaluation rather than reasoning. Common cases are: (1) the model retrieves multiple correct aliases but omits a few, leading to partial correct; and (2) minor formatting differences or string mismatches cause correct answers to be marked as incorrect.



Overall, we observe that E4 (insufficient information) and E2 (parameter errors, particularly in GeneHop) are the main bottlenecks in the current system:

\begin{enumerate}[leftmargin=10pt]
\item \textbf{E4 errors} are common across GeneTuring tasks, especially in Human genome DNA alignment and Gene location. These reflect a fundamental limitation: even with correct reasoning and tool use, the available data may lack sufficient coverage to fully answer the question, indicating an external bottleneck in database quality and scope.
\item \textbf{E2 errors} in GeneHop reveal the model's difficulty with flexible, loosely structured queries that require precise API parameterization. While some failures stem from the zero-shot nature of prompts, ultimately, fine-tuning for better API understanding and parameter selection would provide a more reliable solution.
\end{enumerate}

\subsection{32B Tool Agents underperform 14B ones}

In GeneHop, most errors concentrated in two subtasks: SNP gene function and Disease gene location. These errors were not due to limitations in model capability, but rather differences in reasoning strategies across model sizes.

For Disease gene location, the standard approach involves three steps: (1) retrieving the OMIM ID from the disease name, (2) obtaining related genes, and (3) querying their chromosome locations. The 32B model often attempts shortcuts by directly calling the OMIM esummary to extract gene and location data in one step. While faster, this yields incomplete or inaccurate results. In contrast, the 14B model follows the full pipeline, producing more reliable answers despite longer execution.

A similar pattern appears in SNP gene function tasks. Instead of completing the two-step process (retrieve gene then its function), the 32B model often stops early with vague results, leading to partial scores. The 14B model, unable to find sufficient information in the first step, proceeds to the second, ultimately providing a more complete answer.

These findings highlight that stronger models can underperform when they take overly aggressive shortcuts or terminate reasoning prematurely—illustrating a ``too smart for its own good'' failure mode. This suggests the need for task-specific fine-tuning and stricter control over reasoning chains to ensure consistent performance.

Moreover, we note that GeneGPT's reliance solely on NCBI tools imposes limitations, such as restricted query parameters and lack of relevance-based ranking. For instance, 17\% of GeneTuring's human genome DNA alignment questions remain unanswerable with NCBI alone. Integrating additional data sources and retrieval techniques like query rewriting or expansion could improve coverage and answer quality. 
\vspace{-8pt}
\section{More than Accuracy}

\subsection{Latency analysis} \label{sec_latency}
\begin{figure}[t!]
\centering
\begin{subfigure}[b]{\linewidth}
    \centering
    \includegraphics[width=\linewidth]{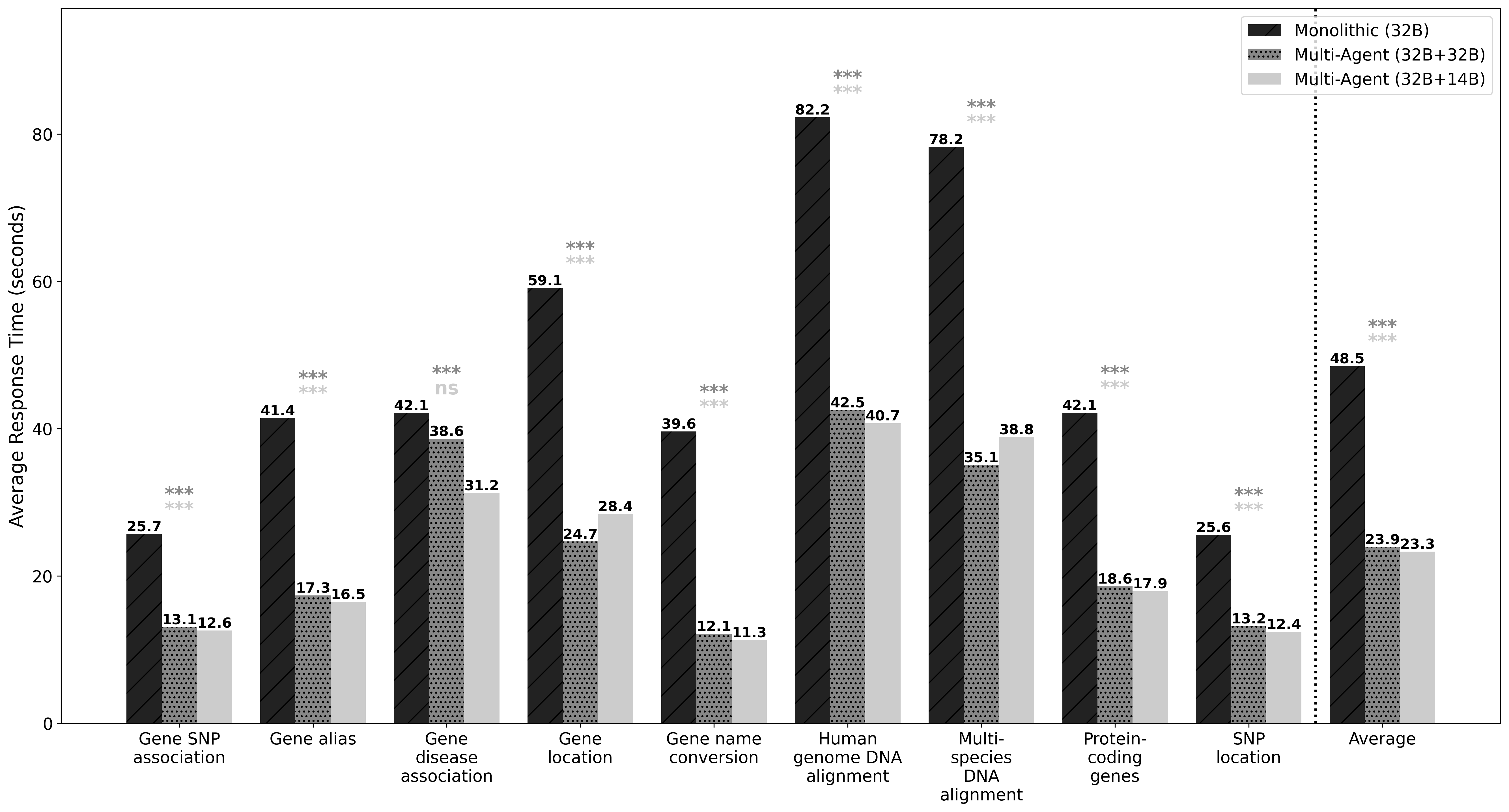}
    \vspace{-0.5cm}
    \caption{GeneTuring}
    \label{fig:geneturing_latency}
\end{subfigure}
\begin{subfigure}[b]{\linewidth}
    \centering
    \includegraphics[width=\linewidth]{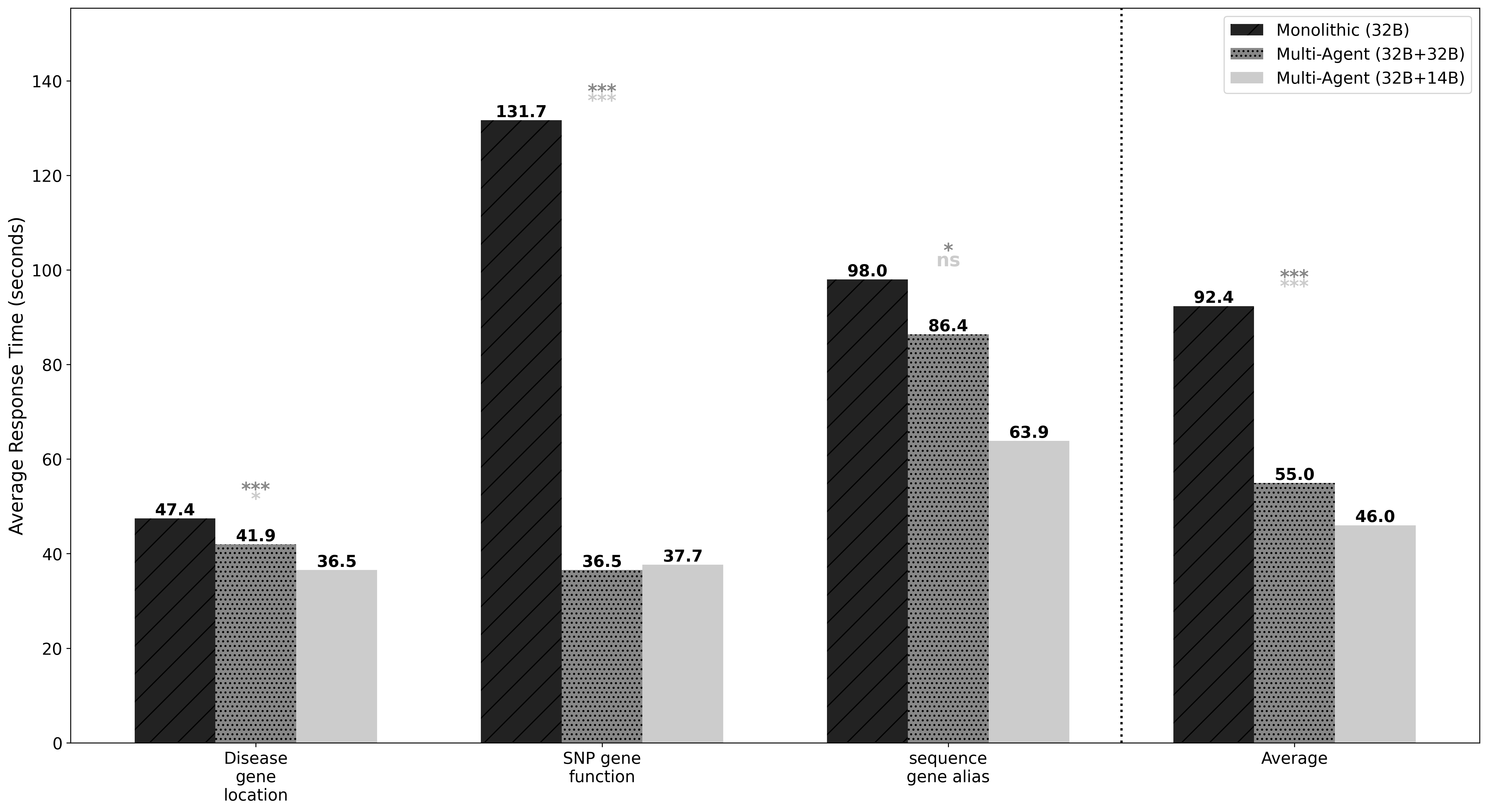}
    \vspace{-0.5cm}
    \caption{GeneHop}
    \label{fig:genehop_latency}
\end{subfigure}

\vspace{-0.2cm}
\caption{Latency per task and overall average latency (in seconds) for the monolithic architecture (Qwen2.5-32B backbone) versus the multi-agent architecture. For the multi-agent setup, we evaluate two model size combinations: 32B+32B and 32B+14B. Statistical significance is reported relative to the monolithic baseline: *** p<0.001, ** p<0.01, * p<0.05; n.s. indicates non-significant differences.}
\label{fig:latency_comparison}
\vspace{-0.75cm}
\end{figure}

To further investigate the system-level benefits of OpenBioLLM's multi-agent design, we compared response times between the monolithic Qwen2.5-32B model and our multi-agent OpenBioLLM framework configured with the same 32B base model for the controller and tool agents (32B+32B). This isolates the impact of architectural differences from that of model capacity. The results are shown in Figures~\ref{fig:geneturing_latency} and~\ref{fig:genehop_latency}, which report per-task latency across GeneTuring and GeneHop benchmarks, respectively.

Across both benchmarks, our multi-agent architecture consistently outperforms the monolithic baseline in terms of response time. In the GeneTuring dataset (Figure~\ref{fig:geneturing_latency}), significant latency reductions are observed in tasks such as Gene alias, Gene location, and Protein-coding genes, with improvements often exceeding 30–40\% ($p~<~0.001$). Gains are largely attributable to the modular execution strategy. Although the overall model size remains the same, each agent handles a narrower subtask with shorter prompts and limited context, which in turn reduces individual model response time and improves overall pipeline efficiency.

In GeneHop (Figure~\ref{fig:genehop_latency}), the benefits are even more pronounced. The most dramatic example is SNP gene function, where the monolithic model takes over 130 seconds on average, while the multi-agent pipeline reduces it to just 36.5 seconds. This illustrates how task decomposition not only enhances reasoning accuracy but also leads to substantial speedups by avoiding overloading a single model with multi-step reasoning and large intermediate context.

Although a few tasks (e.g., Gene disease association, Sequence gene alias) show non-significant differences, the overall trend confirms that modular multi-agent architectures can achieve lower latency even when using the same underlying model capacity. This supports our claim that architectural design plays a critical role in the efficiency of LLM-based systems.

\vspace{-7pt}
\subsection{Other Advantages}

\noindent\textbf{Model Allocation:}
By decomposing the reasoning workflow into specialized sub-tasks, OpenBioLLM aligns model strengths with task complexity. High-capacity models (e.g., 32B) handle complex reasoning, while lighter models (e.g., 14B) process structured queries like parameter extraction and API formatting. This separation improves efficiency and enables further specialization—for instance, replacing each agent with a model fine-tuned for its role.

\vspace{0.5em}\noindent\textbf{Scalability:}
Unlike monolithic designs that rely on hardcoded, fixed workflows, our architecture supports modular expansion. New tool agents—such as specialized APIs or knowledge bases—can be integrated seamlessly by extending the router's decision space. This makes OpenBioLLM adaptable to new domains or tasks without retraining the entire system.

\vspace{0.5em}\noindent\textbf{Traceability:}
Another major advantage of the multi-agent setup is the ability to trace and debug the decision process. Each agent interaction is explicitly recorded, and using the LangSmith platform, we can inspect intermediate steps, tool invocations, and LLM responses. This visibility facilitates both evaluation and troubleshooting, and significantly improves the reliability of deployment. 
\vspace{-0.2cm}
\section{Conclusion}





This study explored the feasibility of open-source large language models (LLMs) for genomic QA. Building on GeneGPT, we developed \textbf{OpenBioLLM}, a modular multi-agent system that integrates tool augmentation, structured reasoning, and transparent API interaction.


We first reproduced the GeneGPT pipeline using Qwen2.5 and Llama3.1, identifying limitations in prompt design, tool invocation, and output formatting. By optimizing prompts and structuring tool integration, we improved performance on both GeneTuring and GeneHop, with Qwen2.5-72B achieving average scores of 0.838 and 0.727 respectively—surpassing the original Codex-based system.


To reduce token overhead and improve interpretability, we adopted a modular architecture with specialized roles (controller, tool agents, evaluator). A 32B+14B setup outperformed larger monolithic models, scoring 0.844 on GeneTuring and 0.830 on GeneHop—showing that smaller, role-specialized models can match or exceed larger ones through task decomposition.



Error analysis showed that most failures were due to missing evidence or parameter errors rather than model limitations. Latency evaluations confirmed the architecture’s efficiency, reducing average response time by ~51\% on GeneTuring and ~40\% on GeneHop compared to monolithic setups. 

\new{We also found that larger tool agents (32B) sometimes underperformed smaller ones (14B) on multi-hop tasks by taking shortcuts and ending reasoning too early. This effect highlights that role-fitness and task-compliance can outweigh model size, which may otherwise backfire in complex reasoning. These insights can extend beyond genomics and offer broader guidance for constructing effective multi-agent systems.}



Overall, OpenBioLLM presents a scalable and interpretable approach for applying open-source LLMs to genomics. We hope this work bridges the gap between general-purpose LLMs and the specialized needs of biomedical research.

\bibliographystyle{ACM-Reference-Format}
\bibliography{GeneQwen.bib}

\end{document}